\documentclass[twocolumn]{svjour3}          
\smartqed  
\usepackage{amsfonts}
\usepackage{amssymb}
\usepackage{graphicx}
\usepackage{subfigure}
\usepackage{color}
\usepackage[cmex10]{amsmath}
\usepackage[boxed,algoruled,linesnumbered]{algorithm2e}
\usepackage[pagebackref=true,breaklinks=true,letterpaper=true,colorlinks,bookmarks=false]{hyperref}
\journalname{Machine Vision and Applications}

\begin{document}

\title{A Multi-Camera Image Processing and Visualization System for Train Safety Assessment}


\author{Giuseppe Lisanti \and
 Svebor Karaman     \and
 Daniele Pezzatini \and
         Alberto Del Bimbo 
}

\institute{G. Lisanti, S. Karaman, D. Pezzatini, A. Del Bimbo \at 
	     Media Integration and Communication Center (MICC),
	      University of Florence, Florence, Italy.\\
              \email\{giuseppe.lisanti, svebor.karaman, daniele.pezzatini,
               alberto.delbimbo\}@unifi.it           
\and
S. Karaman \at
Digital Video and Multimedia Lab (DVMM Lab), Department of Electrical Engineering, Columbia University, New~York, USA\\
\email{svebor.karaman@columbia.edu}
}

\date{Received: date / Accepted: date}

\maketitle

\begin{abstract}
In this paper we present a machine vision system to efficiently monitor, analyze
 and present visual data acquired with a 
railway overhead gantry equipped with multiple cameras. This solution 
aims to
improve the safety of daily life railway transportation in a two-fold
manner: (1) by providing automatic algorithms that can process large
imagery of trains (2) by helping train operators to keep attention on
any possible malfunction. The system is designed with the latest
cutting edge, high-rate visible and thermal cameras that observe a
train passing under an railway overhead gantry. 
The machine vision system is composed of
three principal modules: (1) an automatic wagon
identification system, recognizing the wagon ID according to 
the UIC classification of railway coaches;
 (2) a temperature monitoring system;
(3) a system for the detection, localization and visualization of the pantograph
of the train. 
These three machine vision modules process batch
trains sequences and their resulting analysis are presented to 
an operator using a multitouch user interface.

We detail all technical aspects of our multi-camera portal: the hardware requirements,
the software developed to deal with the high-frame rate cameras and ensure reliable acquisition, 
the algorithms proposed to solve each computer vision task, and the multitouch interaction and visualization interface.
We evaluate each component of our system on a dataset recorded in an ad-hoc railway test-bed, showing the potential of our proposed portal for train safety assessment. 


\end{abstract}

\section{Introduction}
In the last years train safety got the attention of media
and public opinion after several disastrous train accidents, as those
that happened in Italy in 2009~\cite{ViareggioIncident} and in France in 2013~\cite{FrenchTrainCrash}.
Train accidents may be a result of either a problem on the railway tracks, as it was the case for the French accident, or some issues with the train itself.
The analysis of the railway tracks requires the installation of sensors on board of a train that should travel on the tracks that have to be inspected.
Several proposals have been made in this sense~\cite{camargo2011emerging,edwards2009advancements}. 
This work focuses on the safety assessment of the train itself. 

A train is composed of a locomotive and multiple wagons, any of its components can be a risk for the train safety. 
A single wagon failure can trigger the derailment of several wagons and have dramatic consequences. 
The Viareggio accident~\cite{ViareggioIncident} is believed to be the consequences of an axle failure on a tank wagon, the wagon hit the platform of the station and overturned to the left and several following wagons also overturned, exploded and caught fire.
Another important aspect of train safety is temperature monitoring especially when the train is approaching a tunnel where escape in case of fire
can be difficult. For example, the Kaprun disaster~\cite{schupfer2001fire} was due to an electric fan heater that caught fire. 
Hence monitoring an abnormal temperature on any part of the train can provide an early notice of an issue and thus prevent its potential dramatic outcome.
A train can be considered safe for transit if all wagons are adapted for the transit on the railway, the locomotive and all wagons 
exhibit nominal temperatures and no out of shape elements are present.
Failure to fulfill any of these requirements may indicate a risk situation.
The analysis of each train status may be done by stopping and analyzing each train in a offtrack location before being allowed to 
travel. This would induce serious delays on the train traffic.
Hence, more interest have been put on portal based system that could be installed on some important keypoints of the railway network 
(e.g. before a tunnel or before entering a train station) in order to asses on-the-fly all the safety requirements. 
This solution has the clear advantage of not requiring to stop the train to run the analysis and it is also possible to install 
multiple sensors on a single portal providing a thorough analysis of the train status at once. 
However, such portal based approaches require the monitoring system to be able to capture 
all required signals even for a train running at full speed.

This paper depicts our proposed multi-camera portal for train safety assessment developed in the context of the Integrated Intermodal System for Security and Signaling on Rail (SISSI) project, funded by Regione Toscana (Italy). Our system relies on high-speed and thermal cameras
to monitor several aspects of the train. The acquired signals are processed by computer vision methods to extract meaningful information.
Finally, all the information is provided to an operator through a touch-based user interface. 
We first review in the next section the state-of-the-art of computer vision based system for train safety and of touch-based interaction for control rooms and train safety.
We then give an in-depth presentation of our proposed system in section~\ref{sec:oursystem}, specifying the hardware and giving an overview of the software developed to obtain reliable data acquisition from all sensors. In section~\ref{sec:train_analysis}, we detail how we solve each target task of the train analysis,namely the automatic wagon
identification, the temperature monitoring, and the detection and localization of the pantograph
of the train.  
We then describe how all the results are provided to the operator on our multitouch interface. 
Finally in section~\ref{sec:evaluation}, we give an evaluation of each sub-system of our multi-camera portal on a dataset recorded in an ad-hoc railway test-bed.

\section{Related work}
\subsection{Computer vision based systems for train safety}

We can distinguish in the literature the approaches that target safety assessment of the train surroundings or the train itself. Some approaches focus on a single aspect of train safety, while multi-modal portals tries to analyze at the same times multiple safety features.

Many railway accidents happen at railway crossing where an object such as a car is stopped on the railway, hence one common use of computer vision is to detect if an obstacle is obstructing the railway. Machine vision was used in~\cite{pu2014study} to detect moving obstacle in these locations. A 3D vision system for obstacle detection is proposed in~\cite{weichselbaum2013accurate}. Train stations are also a risk environment, in~\cite{delgado2014automatic} the authors present a method for automatically detecting people jumping or falling off a train platform.

Another aspect to consider to assess train safety is the proper configuration of the train itself.
In~\cite{sacchipavisys}, a system to detect misalignment of a train pantograph is proposed.
The authors of~\cite{fumagallimultifunction} proposed a multi-function portal similar in spirit to ours with the main objectives of
detecting misalignment of carriage or abnormal temperature so as to be able to stop a train before it enters a tunnel.
They rely on line-scan cameras to obtain the train image in the visible domain, pyroelectric line cameras for thermal imaging 
and a distributed time-of-flight telemeter for the train shape analysis. The evaluation targets mostly the sensors performance and only qualitative results of out-of-shape detections are given.

\begin{table*}[t]
\begin{center}
\scalebox{0.85}{
\renewcommand{\arraystretch}{1.2}
\begin{tabular}{ |c|c|c|c|c|c|c|c|c| }
 \hline
 \textbf{Camera Model} &  \textbf{Producer} & \textbf{Data type} & \textbf{Matrix/Line} &   \textbf{Hz} &  \textbf{Resolution} &  \textbf{Data rate} &\textbf{Temperature range} & \textbf{\# of Sensors}\\
  \hline
  HM-640 & Teledyne Dalsa & Visual & Matrix & 300 & 640x480 & 92MB/s  & -- & 1\\
    \hline
    Spyder 4K  & Teledyne Dalsa & Visual &  Line & 18500 & 4096x1 & 80MB/s  & -- & 2\\
      \hline
      256L & PYROLINE & Thermal & Line & 512 & 256x1 & 128KB/s & $[30^{\circ} ... 800^{\circ}]$ & 2\\
  \hline
\end{tabular}
}
\end{center}
\caption{All our sensors references, characteristics and count.}
\label{table:sensors}
\end{table*}

\subsection{Touch-based interface for control rooms and train safety}
Operators in control rooms are often asked to monitor multiple safety characteristics and have to perform crucial security operations in a short amount of time. This is the main reason why information visualization and touch-based interaction play a key role when developing a monitoring tool for a control room. 
Several studies~\cite{kin2009determining,Forlines:2007:DVM:1240624.1240726} have been done to assess the benefits of the adoption of a multitouch workstation for tasks that require interaction with multiple visual cues. Results have shown that multitouch interaction can be twice as fast as a mouse based one. Furthermore, multitouch interaction is often preferred by the users due to the direct manipulation of graphical elements offered, resulting in a more natural and effective approach to carry out the requested tasks.
Touch-based interaction has been exploited in control and security process since the early seventies. In 1973, Beck and Stumpe~\cite{beck1973two} proposed a prototype of touchscreen device to control the new CERN accelerator. In recent years, studies were conducted to propose and evaluate good practices in the design process of touch-based interfaces for security operators. Zahler~\cite{zahler2008design} proposes multiple patterns for the design of touch-based user interface for railways security and other safety-critical applications.
In~\cite{bjorneseth2012assessing}, the authors investigate the effectiveness of direct manipulation in multitouch interfaces for safety-critical situations in maritime control room. Results showed that direct manipulation of interface elements can enhance situational awareness of users.

Evaluation and testing safety-critical interfaces is crucial to show whether a novel developed system actually fulfills its goals.
Authors of~\cite{stelzer2014evaluating} propose a method for the evaluation of user interface for safety in railway based on a high-fidelity simulator of an interlocking systems. 
Although many standardized usability evaluation methods exist and are commonly used for general purpose systems, some specific methods have been defined for the evaluation of safety critical interactive systems~\cite{thimbleby2007interaction}.

\subsection{Contribution}

Our proposal is to use high-speed cameras mounted on a railway overhead gantry to monitor multiple aspects of the train. 
In particular, we have designed a machine vision system that coordinate different sensors with different speed to: (1) automatically segment the  wagon identifier according to the UIC classification of railway coaches;
	(2) extract the wagon temperature to prevent fires and flames on board; and 
	(3) detect the pantograph passage.
	
We propose a touch-based interface that adopts interaction metaphors like direct manipulation and multitouch gestures. The goal of the touch-based user interface is to give operators a quick and efficient way to interact with results of the video sequences analysis. Manipulating all the output of the machine vision system, the operator is able to efficiently control the train's safety requirements.

\begin{figure}[b]
\centering
\includegraphics[width=1\columnwidth]{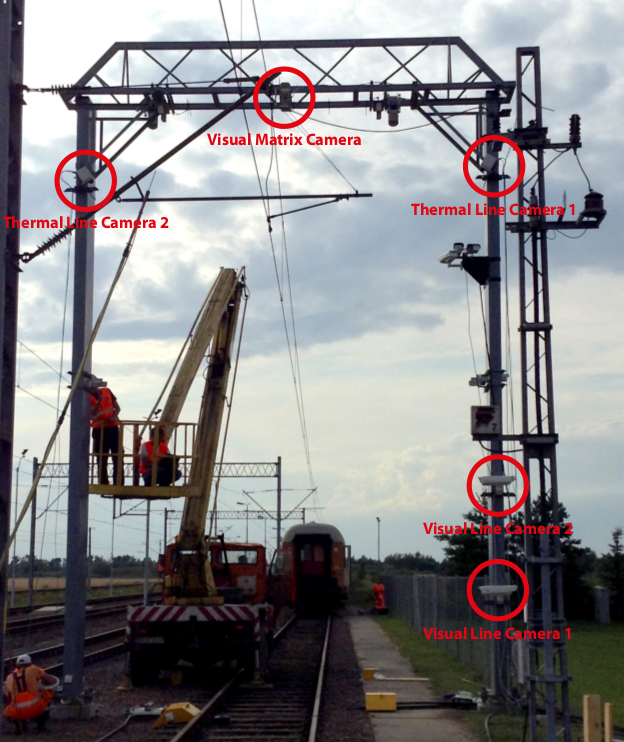}
\caption{The portal equipped with all cameras. The visual matrix camera (HM-640) is on top in the center of the gantry, two thermal cameras (256L) are positioned on each side of the portal, and two visual line cameras (Spyder 4K) are positioned on the same side but one on top of the other.}
\label{fig:portal}
\end{figure}

\section{Our multi-sensors portal}
\label{sec:oursystem}

In this section we describe the physical structure and the sensors characteristics of our multi-camera portal. We then detail the acquisition manager we designed to manage all the sensors together and deal with their high-rate acquisition.

\begin{figure*}[t]
\centering
\includegraphics[width=1\textwidth]{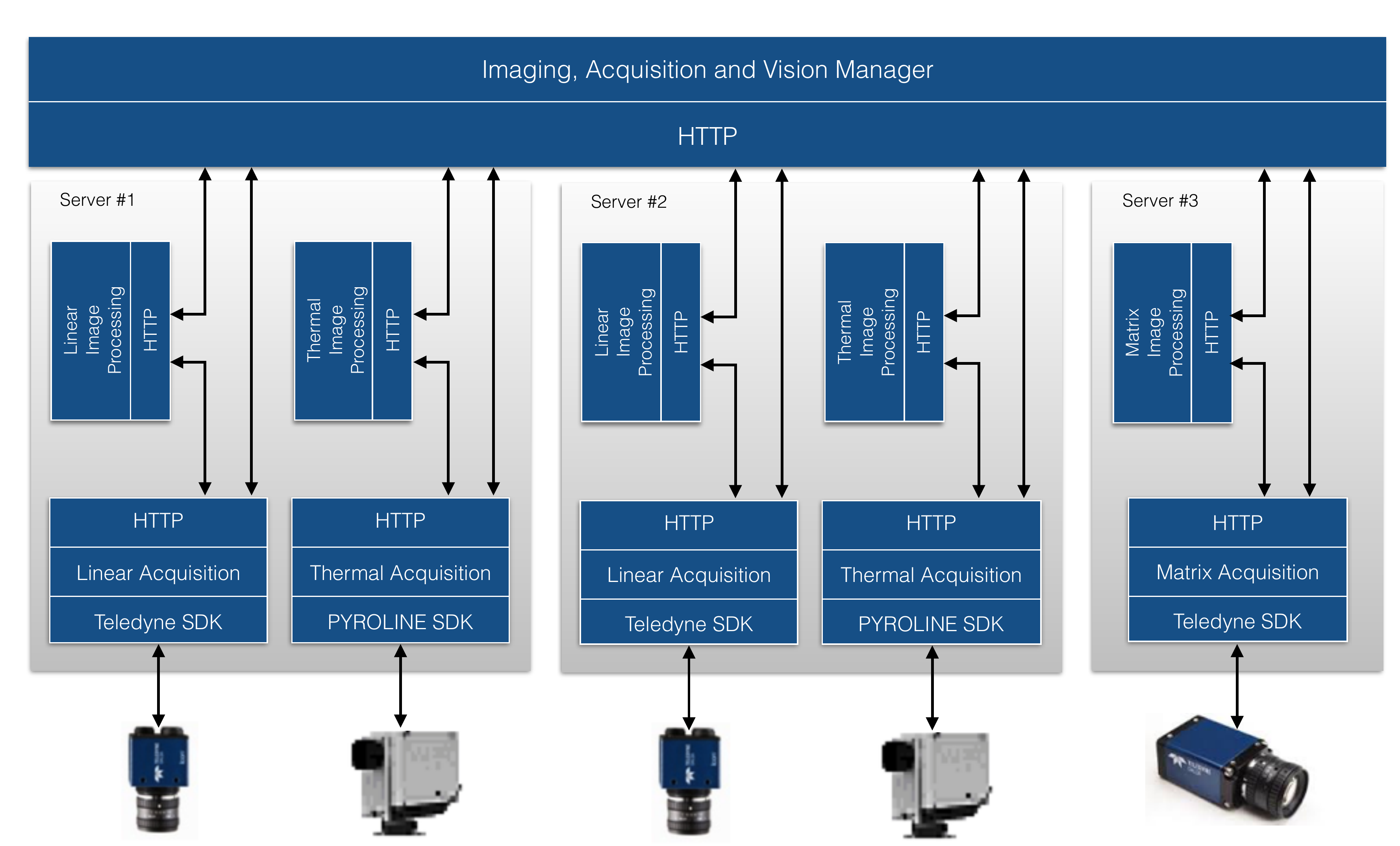}
\caption{Acquisition and processing system architecture}
\label{fig:acqsystem}
\end{figure*}

\subsection{Architecture of the portal and sensors involved}
The portal is built over a single rail, has a height of 8.5 meters and is 6 meters large.  The distance from the train side is around 1.5 meters. 
This gantry is equipped with a total of 5 sensors, the \figurename~\ref{fig:portal} illustrate our portal configuration.

We used three different types of high rate sensors. We summarize the characteristics of each sensor in Table~\ref{table:sensors}. Two of this sensors work in the visual spectrum (one linear and the other matricial) 
while the last one works in the thermal spectrum (with linear acquisition).

In particular, the matrix camera operating in the visual spectrum (Teledyne Dalsa HM-640) acquire 300 grayscale images per second at a resolution of 640x480 pixels. This camera is positioned on the top and at the center of the portal and is used as a general overview of the train passing by and for the visual analysis of the state of the pantograph. 

The linear camera operating in the visual spectrum (Teledyne Dalsa Spyder 4K) acquires 18500 grayscale lines per second with a height of 4096 pixels. We have positioned two linear cameras on the side of the portal: one is used to observe the bottom and central part of the train and is the entry signal of the wagon identification system, see section~\ref{sec:wagonID}; the second linear camera is positioned higher on the side of the portal to capture the top of the train and it will be used for the detection of the pantograph detailed in section~\ref{sec:pantograph}. 

Finally, the linear camera operating in the thermal spectrum (PYROLINE 256L) acquires 512 lines of 256px each with a temperature range of $[30^{\circ} ... 800^{\circ}]$. One of this thermal sensor is positioned on the top of each side of the portal, and they are used to monitor the temperature anywhere in the passing train as explained in section~\ref{sec:temp}.

\subsection{Acquisition framework}

Full synchronization of all the sensors can be really difficult due to the high and different rates of acquisition. However, each sensor being devoted to a specific function, we don't need full and perfect synchronization of the acquisition. We designed a specific hardware/software solution that allows us to obtain a coarse synchronization between all the sensors, enabling a meaningful and easily interpretable playback of the acquisition for the operator. 

As regards hardware, we designed three separate servers to deal with the large amount of data induced by the high-rate cameras mounted on the portal. Specifically, there are 2 servers acquiring data from one linear and one thermal camera, while the third server deals only with the matricial camera as shown in the overview of our system architecture given in \figurename~\ref{fig:acqsystem}. These servers have 8 SAS disk in RAID-01 to obtain the sufficient speed needed (about 270MB/s) to write all the data generated, while maintaining sufficient reliability. 

Concerning the software, we designed an effective solution that allows us to contemporary control each sensor focusing on the optimization of CPU, memory and disk usage. 
In particular, we designed a frame-grabber for each type of sensor: matrix, linear, visual or thermal; exploiting the respective SDK given by the cameras vendor. 

As shown in \figurename~\ref{fig:acqsystem} each frame grabber is controlled by a HTTP server (AcquisitionServer).
For each camera this server implements some acquisition primitives (e.g. start, stop, pause) as well as some specific function depending on the camera model, for example, focus control for the thermal cameras. 

To contemporary control each AcquisitionServer (of each sensor) we designed another HTTP server called the AcquisitionManager. Each AcquisitionServer registers to the AcquisitionManager and periodically send its state. Once an AcquisitionServer is registered to the AcquisitionManager it is possible, using a simple web interface we developed, to control the IP address, the state of the grabber, and all the primitives expected for the relative sensor.  

The acquisition primitives in common between all sensors are shown as a unique button in the web interface of the AcquisitionManager, while the primitives specific to a sensor are shown only for the registered sensor that can use them. 
Having a common interface showing the state of all sensors is particularly useful, for example, to prevent starting a new acquisition or stopping an ongoing acquisition if some of the AcquisitionServers is still saving some recently acquired data.
When an AcquisitionServer is closed it unregisters from the AcquisitionManager. 
This software design offers the advantage that a sensor can be easily added, activated or deactivated for a specific acquisition.

\section{Train analysis}
\label{sec:train_analysis}

The aim of the proposed system is to analyze both the visual and thermal data extracted using the high-rate sensors, described in the previous section. In particular, we developed three sub-systems in order to recognize the wagon identifier, monitor the temperature, and detect the pantograph. Finally, we will give an overview of the multitouch user interface we designed to enable an operator to interact with the processing results in the control room.

\subsection{Wagon identification}
\label{sec:wagonID}
The \emph{Wagon identifier subsystem} aims to identify the wagon by segmenting its unique international identification number from the image acquired with the Visual Line Camera 1 positioned at the bottom right of the portal, see \figurename~\ref{fig:portal}. From this identifier multiple characteristics can be extracted (type of wagon/locomotive, owner and country for example) and thus one can understand if the wagon is expected and allowed to transit on the monitored railway section. Due to the huge dimension of the image and the presence of noise we need to apply a robust identifier segmentation method. The whole method, described in Algorithm~\ref{alg:text}, relies on image processing and geometric analysis to obtain the position of the identification number in the image. 

\begin{algorithm}[t]
\KwIn{$\mathbf{I}$, $r_D$, $d$, $s$, $w$, $h$}
\KwOut{$\mathbf{\widehat{b}}$}
\vspace{0.2cm}
Compute $\tau_O = AdaptiveThreshold(\mathbf{I})$; \\
Extract edge $\mathbf{I}_e = CannyEdgeDetector(\mathbf{I},\tau_O)$; \\
Perform morphological dilation $\mathbf{I}_D  = (\mathbf{I_e} \oplus disk(r_D)$); \\
Perform hole filling $\mathbf{I}_f = fill(\mathbf{I}_D)$; \\
Extract Connected Components $\mathbf{c}_{bbox} = extractCC(\mathbf{I}_f);$ \\
\vspace{0.2cm}
Initialize votes $\mathbf{v} \leftarrow \mathbf{0}$; \\
Initialize $j \leftarrow 0$; \\
Initialize $k \leftarrow 0$; \\
\While{$j \le w$}{ 
\While{$k \le h$}{ 
$\mathbf{\widehat{c}}_{bbox} = SelectCC(\mathbf{c}_{bbox},j,k,d)$; \\
$\mathbf{in} = RansacFitLine(\mathbf{\widehat{c}}_{bbox})$; \\
$\mathbf{v} = \mathbf{v} + Voting(\mathbf{c}_{bbox}, \mathbf{in}); $ \\
k = k + s; \\
}
j = j + s; \\
}
$\mathbf{\widehat{b}} = SegmentSalientRegions(\mathbf{I},\mathbf{c}_{bbox},\mathbf{v}); $ \\
\caption{Wagon ID segmentation.}
\label{alg:text}
\normalsize
\end{algorithm}

Given an image of a wagon $\mathbf{I}$ of width $w$ and height $h$, we first apply an adaptive thresholding method to find the optimal threshold that separate foreground from background pixels~\cite{1979:ots}, such as:
\begin{align}
\tau_O = AdaptiveThreshold(\mathbf{I})
\end{align}
The Otsu adaptive thresholding algorithm assumes that the image contains two classes of pixels (e.g. foreground and background) then calculates the optimum threshold separating these two classes in order to minimize intra-class variance.
After that, the threshold $\tau_O$ is used as input for the Canny edge detector~\cite{Canny:1986:CAE:11274.11275} to segment the contour of the foreground elements present in the image. 
\begin{equation}
\mathbf{I}_e = CannyEdgeDetector(\mathbf{I},\tau_O).
\end{equation}
The regions defined by connected edges are filled first by using the morphological operation of dilation and then with a fill operation to definitely close small holes:
\begin{eqnarray}
\mathbf{I}_D  & = & (\mathbf{I_e} \oplus disk(r_D)),\\
\mathbf{I}_f & = & fill(\mathbf{I}_D),
\end{eqnarray}
where $r_D$ represent the disk ray size used for the dilation operation.
Once those regions are filled, a connected components labelling algorithm is used to define the bounding boxes containing the blob regions previously segmented:
\begin{equation}
\centering
\mathbf{c}_{bbox} = extractCC(\mathbf{I}_f).
\end{equation}
Since the connected components generally correspond to the foreground objects in the image, we can say that after the labelling we are able to know how many foreground objects are contained in the image and what are the pixels that belong to each object. The question remaining to solve is which of these objects are characters of the wagon identifier.

In order to identify the connected components corresponding to characters we apply a voting procedure based on the sliding-window paradigm. In particular, given all the bounding boxes of the connected components we can infer that the characters of the identifier are close to each other and mostly aligned along a line. 
For this purpose, we apply a sliding-window procedure (with sampling step of $s$ pixels) to the image and for each sub-window, of dimension $d\times d$ pixels, we estimate a line through the RANSAC algorithm~\cite{Fischler:1981:RSC:358669.358692} considering only the bottom-right points of the bounding boxes present in that sub-windows:
\begin{eqnarray}
\mathbf{\widehat{c}}_{bbox} & = & SelectCC(\mathbf{c}_{bbox},j,k,d),\\
\mathbf{in} & = & RansacFitLine(\mathbf{\widehat{c}}_{bbox}),
\end{eqnarray}
where $j$ and $k$ represent the top right coordinate of the sub-windows considered. In each sliding-window, the points $\mathbf{in}$ selected by RANSAC as inliers accumulate a vote. At the end of this procedure the points with the most votes will represent the bounding box with a higher probability of containing a character:
\begin{equation}
\mathbf{v} = \mathbf{v} + Voting(\mathbf{c}_{bbox}, \mathbf{in}).
\end{equation}

\begin{figure}
\centering
\includegraphics[width=\columnwidth]{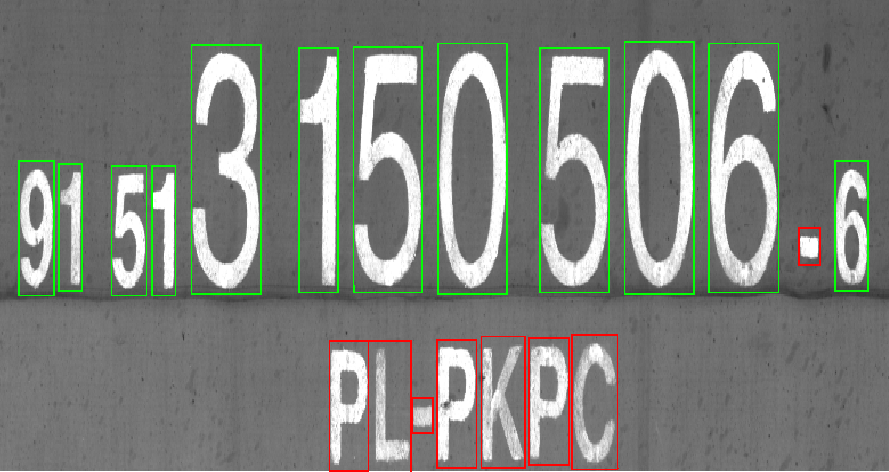}
\caption{Example of wagon id segmentation.}
\label{fig:id_sample}
\end{figure}

Finally, to obtain the identifier which is composed of 12 characters we selected the sub-region of the image $\mathbf{\widehat{b}}$ containing the most voted and aligned foreground objects. The alignment is estimated by computing the distances on the x-axis $D_x$ and y-axis $D_y$ for the 20 most important foreground regions according to $\mathbf{v}$. Then we take the exponential of the negative of these distances and we weight the votes previously obtained with those matrices separately.
\begin{equation}
\mathbf{v}_w = \texttt{exp}(-D_x)*\mathbf{v}+\texttt{exp}(-D_y)*\mathbf{v}.
\end{equation}
All those regions with a weighted vote $\mathbf{v}_w$ greater than zero represents a character of the ID. We then take a crop of the original image as the region containing the set of ID characters, see the example in \figurename~\ref{fig:id_sample}. From this image, any Optical Character Recognition (OCR)  method can be applied to obtain the identifier. This identifier segmentation step is necessary as wagon image have an average size of $4096 \times 80000$ and cannot be processed as is by an OCR.

\subsection{Temperatures segmentation}
\label{sec:temp}
The \emph{Thermal monitoring subsystem} acquires two thermal maps of each wagon and compare them to nominal operating temperatures in order to issue an alarm in case of fire risk due to abnormally high temperatures. The sensors involved in this subsystem are the Thermal Line Camera 1 and 2 positioned at the top right and left of the portal, see \figurename~\ref{fig:portal}. 

\begin{figure}[b]
\centering
\subfigure{\includegraphics[width=\columnwidth]{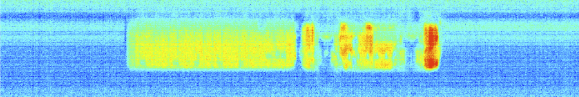}}
\subfigure{\includegraphics[width=\columnwidth]{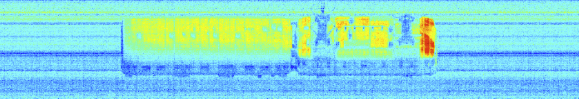}}
\caption{Example of thermal images acquired by the two cameras. The hottest part (in red) is the locomotive engine.}
\label{fig:thermal_samples}
\end{figure}

Each thermal camera is connected and managed by a different server in order to ensure a higher robustness of the thermal subsystem through duplication.  The mosaic image obtained from all the acquired lines concatenation, see examples in \figurename~\ref{fig:thermal_samples}, is divided into subregions of fixed size and for each subregion both the mean and maximum temperatures are calculated. 
The minimum and maximum temperatures coming from one camera are compared with those extracted from the other camera in order to validate their output and ensure that both the servers and sensors are working correctly.

An important phenomenon to be considered is the distortion of the temperature values caused by the perspective in the image. In particular, the pixels furthest from the center of the sensor will be subject to a high distortion caused by the perspective between the sensor and the observed wagon (obviously this phenomenon depends also on the distance from the sensor to the wagon). For this reason we used two cameras observing the wagon from two different viewpoints. 

\subsection{Pantograph detection}
\label{sec:pantograph}
The \emph{Pantograph detection subsystem} detects the passage of the pantograph in order to avoid false positive cuts in a laser based system\footnote{This system is composed of three infrared laser mounted on the portal. This proprietary solution was developed by Thales Italia and cannot be discussed in the scope of this paper.} that analyze the shape of each wagon. The pantograph detection is run on the data extracted from the Visual Line Camera 2 positioned at the middle right of the portal. The obtained segmented high resolution image can also be analyzed by an operator to determine if there is any anomaly in the pantograph shape.

\begin{figure}
\centering
\includegraphics[width=1\columnwidth]{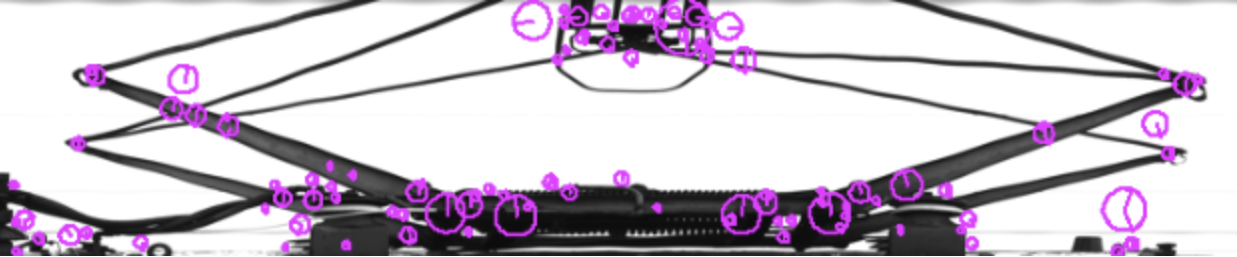}
\caption{SIFT point extracted from the pantograph image template.}
\label{fig:pantograph}
\end{figure}

\noindent The proposed solution is composed by an offline phase and an online phase. Offline, we extract SIFT~\cite{Lowe:2004:DIF:993451.996342} keypoints from an image of the pantograph used as template $\mathbf{T}$, see \figurename~\ref{fig:pantograph}:
\begin{equation}
\mathbf{D}_T = ExtractSIFT(\mathbf{T}).
\end{equation}
The SIFT descriptors are representations of image regions highly discriminative and invariant to changes in brightness, scale and rotations. We store the extracted descriptors in a KD-Tree~\cite{muja_flann_2009} in order to speedup the matching process:
\begin{equation}
\mathbf{K}_\mathbf{T} = KDTree(\mathbf{D}_T).
\end{equation}
Online, once the mosaic image of the train side is obtained, the proposed algorithm~\ref{alg:pantograph} extracts SIFT keypoints from that image:
\begin{equation}
\mathbf{D}_c = ExtractSIFT(\mathbf{I}_c), \\
\end{equation}
 
The KD-Tree nearest neighbor search provides the identifier of the closest descriptor, however, due to the curse of dimensionality, descriptors neighbors in $\mathbb{R}^{128}$, could be not visually similar. For this reason a second filtering is introduced. The distance between the first and the second more similar descriptor is measured and the match are discarded according to $\frac{\mathbf{d}_m^1}{\mathbf{d}_m^2} \leq \tau_d$, where $\tau_d = 0.67$, as in~\cite{Brown03}:
\begin{equation}
\mathbf{D}_m = MatchDescriptors(\mathbf{T},\mathbf{D}_c,\tau_d).
\end{equation}
However, these matches are not guaranteed to be correct, this can occur for various reasons, for example repeated structures in the image or points with similar SIFT descriptors. For this reason a third validation step is performed by applying a geometric robust validation following a projective model transformation. This is obtained  by exploiting the  RANSAC algorithm~\cite{Fischler:1981:RSC:358669.358692} to 
fit a projective model and successively by applying a consistency check algorithm to the estimated homography in order to determine if the fitted model is correct:

\begin{eqnarray}
[\mathbf{H},~\mathbf{in}] & = & RansacFitProj(\mathbf{D}_m);\\
\mathbf{p}_{bbox} & = & CheckGeomConsistency(\mathbf{H},\mathbf{in}).
\end{eqnarray}
In this way it is possible to establish the presence of the pantograph and it is also possible to have an indication of its location in the image and segment the relative sub-image to be shown to the operator.

\begin{algorithm}[t]
\KwIn{$\mathbf{I}_c$, $\mathbf{T}_c$, $\tau_d$}
\KwOut{$\mathbf{p}_{bbox}$}
\vspace{0.2cm}
\textbf{Offline:}\\
$\mathbf{D}_T = ExtractSIFT(\mathbf{T})$; \\
$\mathbf{K}_\mathbf{T} = KDTree(\mathbf{D}_T)$; \\
\vspace{0.2cm}
\textbf{Online:} \\
$\mathbf{D}_c = ExtractSIFT(\mathbf{I}_c)$; \\
$\mathbf{D}_m = MatchDescriptors(\mathbf{T},\mathbf{D}_c,\tau_d)$; \\
$[\mathbf{H},~ \mathbf{in}] = RansacFitProj(\mathbf{D}_m)$;\\
$\mathbf{p}_{bbox} = CheckGeomConsistency(\mathbf{H},\mathbf{in})$;
\caption{Pantograph detection.}
\label{alg:pantograph}
\normalsize
\end{algorithm}

\begin{figure*}
\centering
\subfigure[]{\includegraphics[width=0.562\textwidth]{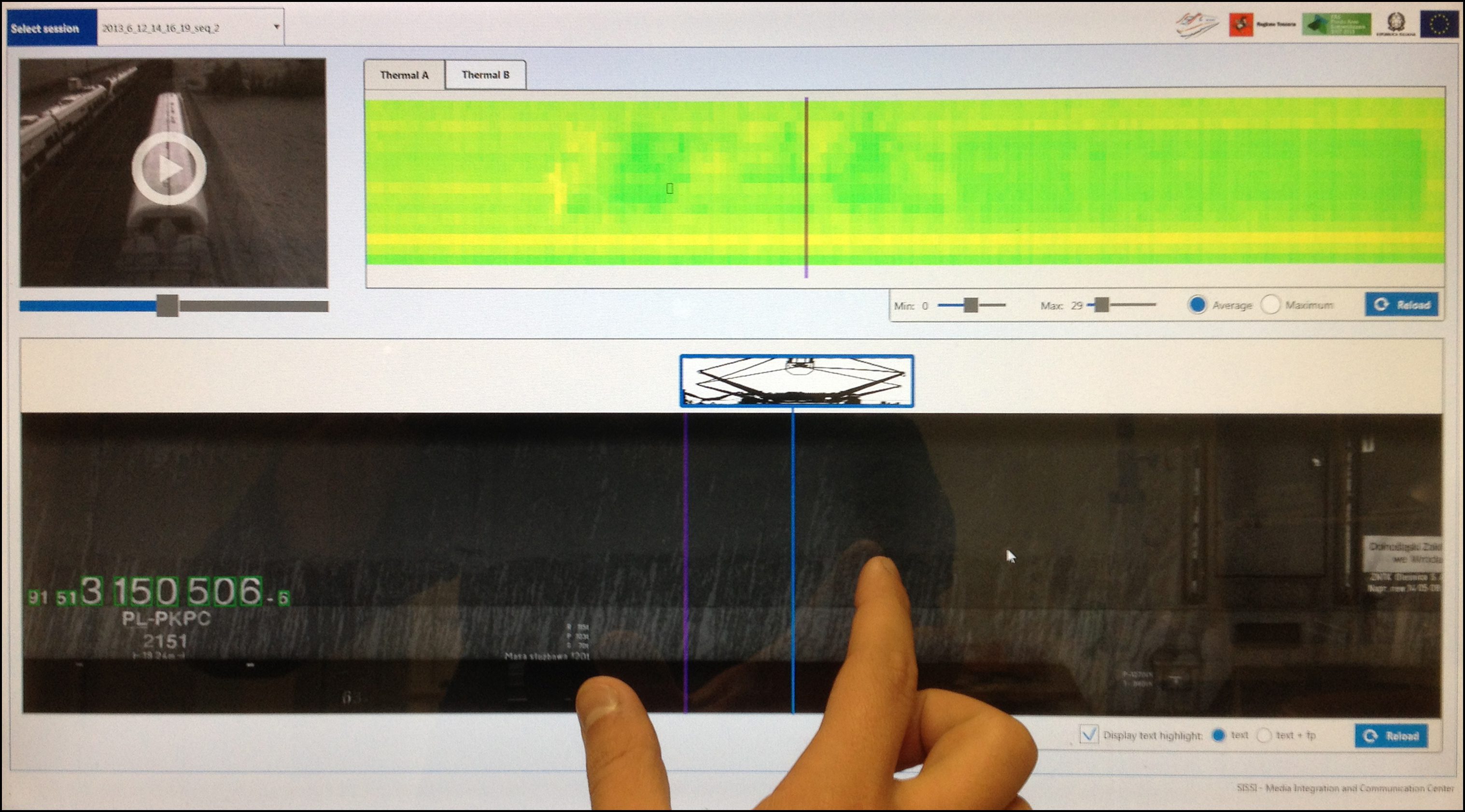}}
\subfigure[]{\includegraphics[width=0.41\textwidth]{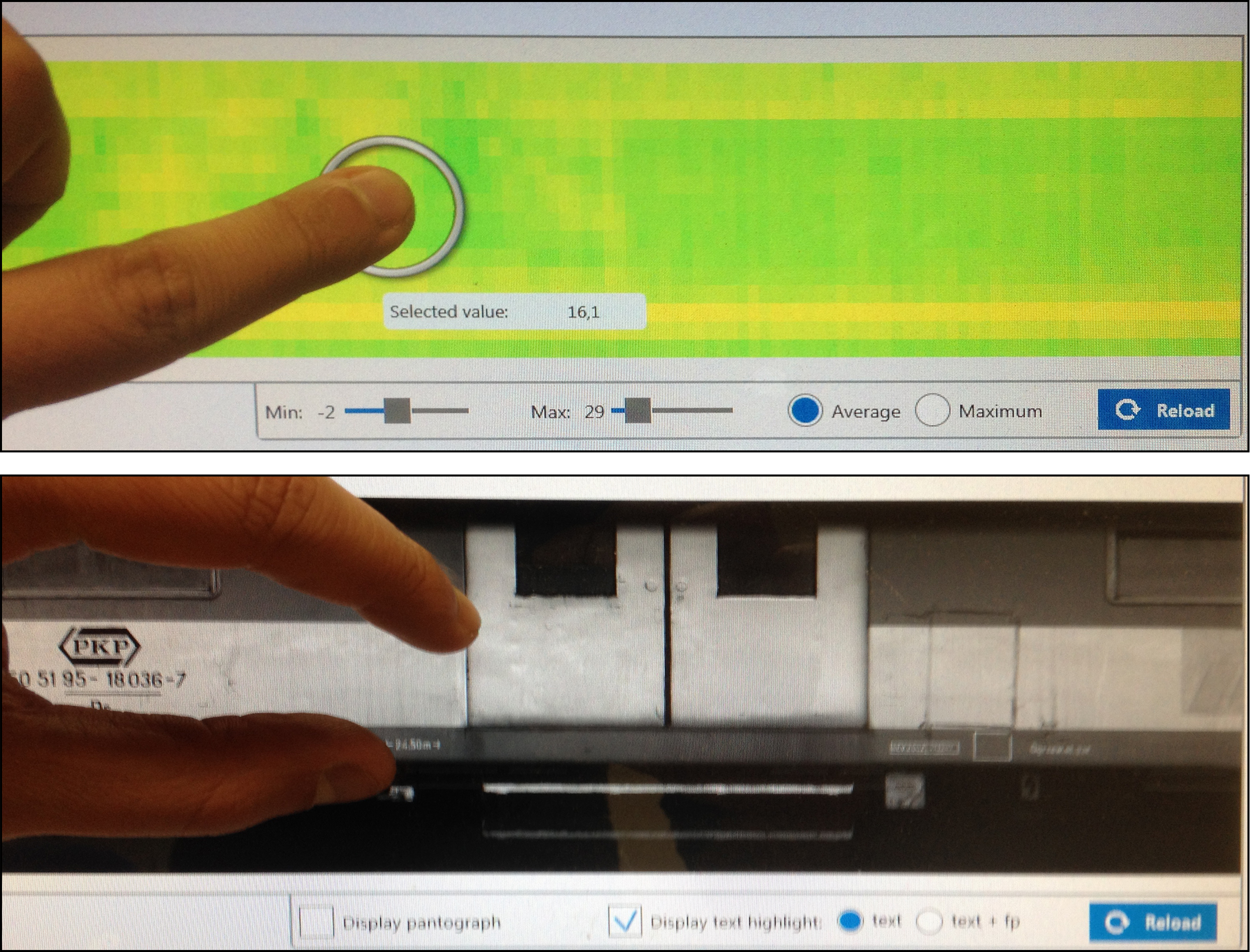}}
\caption{(a) Overview of the interactive interface. (b) Examples of interaction with the multitouch user interface. On top the operator checks punctual temperature value, on the bottom she perform a zoom of an high-definition image of the train}
\label{fig:table_sample}
\end{figure*}

\subsection{Touch-based user interface}

To enable an operator to visualize and interact intuitively with the results of the wagon analysis we developed a touch-based graphical user interface (GUI) based on multitouch interactions. The aim of the GUI is twofold. On one hand, it is used to exhibit all the results of the acquisition and analysis to the security operator in a simple way so he can quickly get an overview of the train status. On the other hand it  provides the operator with several tools for a direct and easy manipulation of all the data necessary to assess the train safety requirements.

An operator first loads a session of a processed wagon analysis in the interface. A session is composed by (i) a frontal video of the train obtained from the matrix camera positioned at the top of our portal, (ii) two thermal images that are the output of the thermal monitoring subsystem, (iii) a high-resolution image of the train acquired by the visual linear camera 1.
The frontal video can be played and scanned through a timeline visualizer. The timeline of the video is synchronized with visual markers on both the thermal and linear imagery in order to give a visual time reference on all results.

Thermal images are displayed with a false-color scale obtained from the temperature values. By default the scale is based on the \textit{max} and the \textit{min} values of the image, but the operator can manually change the range of colors in order to enhance the visualization of specific temperature values. Left and right thermal images can be activated with a selector, so that only one image at a time is visualised in the interface.

The linear camera acquisition result is a high resolution image of the train. In order to allow a fluid and smooth manipulation of this image  we adopted a multi-resolution tiling technique~\cite{y2008methods}. Acquired images are pre-processed in order to have a set of downscaled versions of high-definition ones. Each downscaled version is then decomposed in tiles of 256x256 pixels. The rendering engine of the GUI loads and display only the tiles required for the current zoom level and portion of the image visualized by the operator, instead of loading the entire high-definition image. The operator can activate graphical overlays on the train image in order to visualize the results obtained by the \emph{pantograph detection subsystem} and the \emph{wagon identifier subsystem}. 

\figurename~\ref{fig:table_sample} shows an overview of the interface and some phases of the interaction of a control operator with the interactive GUI, like checking temperature of an area of the wagon or visualising an high-definition image of the train.
The set of functionalities provided by the GUI allows the operator to have a quick overview of the image processing analysis results and to perform punctual and precise controls through direct manipulation using multitouch gestures.

\section{System evaluation}

\label{sec:evaluation}

To evaluate our proposed system we recorded a dataset of sequences using the portal depicted in \figurename~\ref{fig:portal} in Poland (Zmigrod). We acquired 36 sequences of a train composed of one locomotive and one wagon on a test-bed railway track of 1 Km. The train passed under the  multi-camera gantry at different time of the day, at different speeds and with different weather conditions.
To register these sequences we used the system architecture and the web interface previously described in section~\ref{sec:oursystem}.
In this section we will evaluate each of the sub-systems of our proposed approach.

\subsection{Wagon identification}

\begin{figure}
\centering
\includegraphics[width=1\columnwidth]{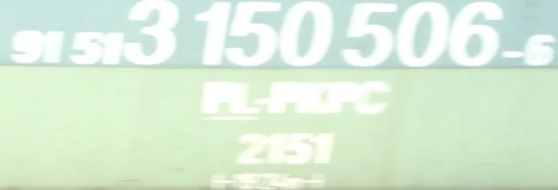}
\caption{Acquisition sample of a full HD video taken with a standard camera.}
\label{fig:bad}
\end{figure}

As shown in \figurename~\ref{fig:bad} it would be really difficult to segment the identifier with a standard camera. Indeed, a high motion blur due to the train speed affects the readability of several characters. The use of a high-rate linear camera is hence necessary to be able to properly segment the wagon identifier.
To evaluate the wagon identifier segmentation performance, we estimated the accuracy of both the full train ID and the single characters segmentation for each wagon. In particular, for the case of characters segmentation we count as true positive every detected region that contains a character of the wagon ID, as false negative every missed character of the wagon ID and as false positive every region classified as part of the ID but non containing a character of the wagon ID. While for the full ID segmentation, we count a true positive every time all the characters of the wagon ID are recognized, a false negative every time at least one character of the wagon ID is missed and as false positive all the regions classified as positive but that do not contain a character of the wagon ID. For the full ID segmentation evaluation, there is exactly one target detection by wagon making it easier to obtain higher false positive rate as any region that do not contains  the ID will be counted as a false positive.

\begin{figure}
	\centering
	\includegraphics[width=0.49\textwidth]{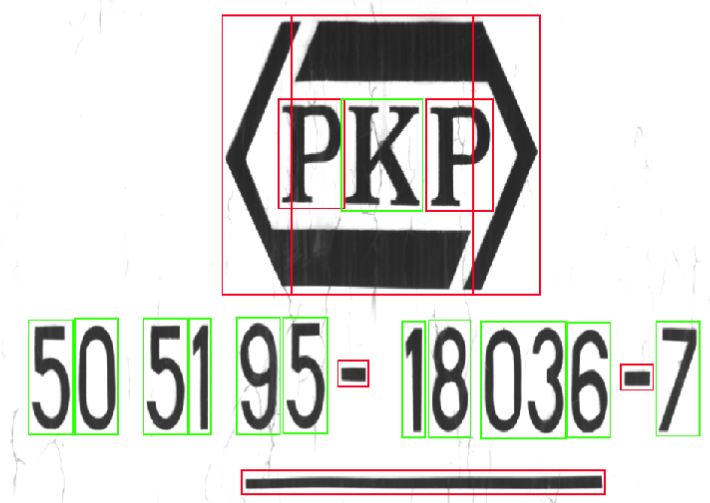}
	\caption{Example of false positive when two character regions (here 0 and 3) are merged.}
	\label{fig:text_fp}
\end{figure}

As it can be observed from Table~\ref{table:textacc_id} the accuracy of the system is very high for full ID segmentation in the case of wagon 1 while for the wagon 2 we are always able to detect the full train ID. 
We can also appreciate that both false positives and false negatives are limited for the full train ID segmentation of each wagon. When evaluating in terms of character segmentation, see Table~\ref{table:textacc}, the results are even better with really low false negative and false positive rates.
False positives are mainly caused by the fact that sometimes small character in the train ID are merged together and a region close to the train ID can be considered part of it, as shown in Fig.~\ref{fig:text_fp}.

\begin{figure*}
\centering
\includegraphics[width=\textwidth]{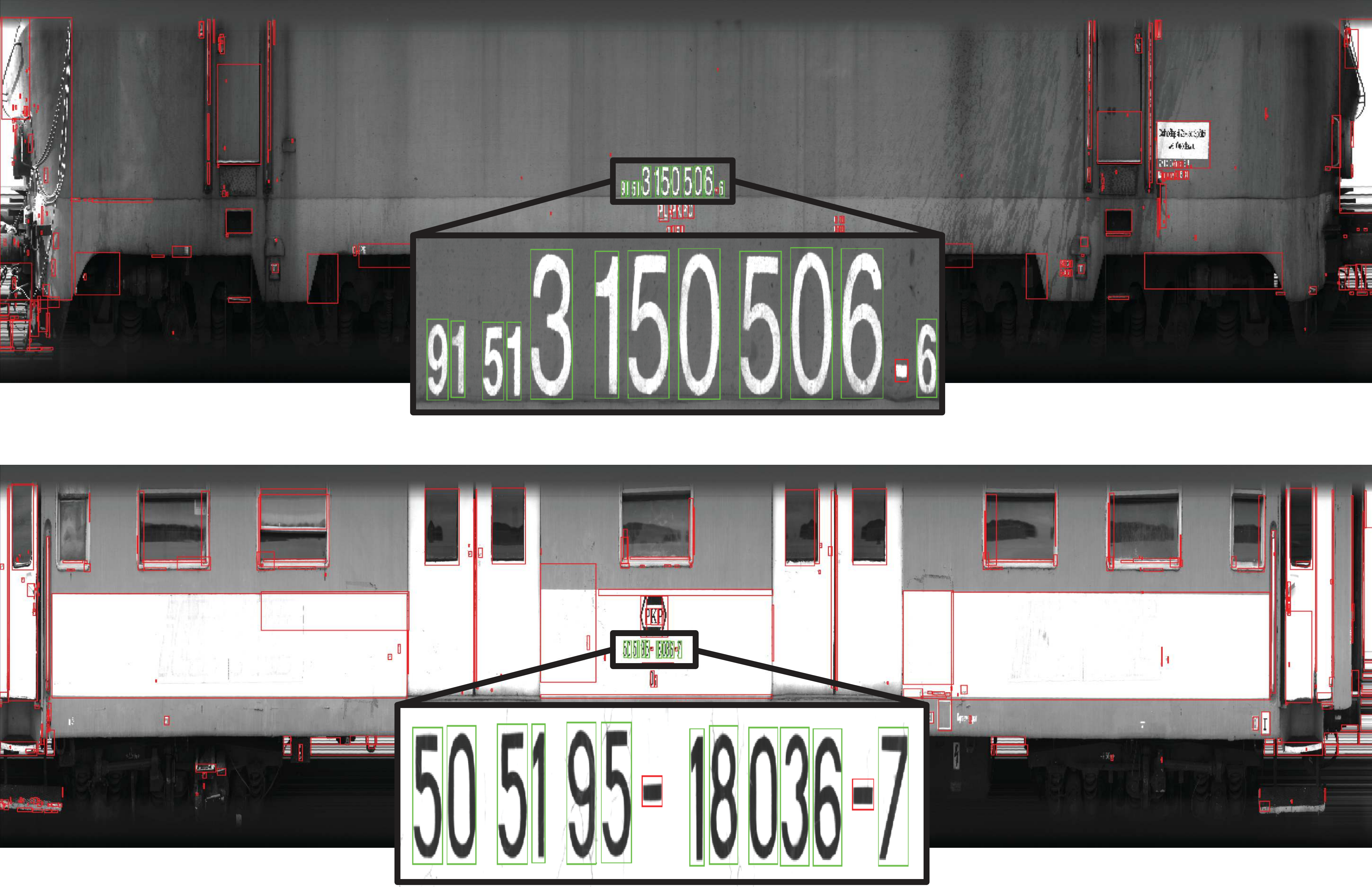}
\caption{Examples of text region segmentation on the two different wagons. In green the bounding box segmented after RANSAC refinement. }
\label{fig:text}
\end{figure*}

In \figurename~\ref{fig:text} we show a qualitative sample of how the proposed solution segment the train identifier, for both the locomotive and the wagon. One can observe how the train identifier is a very small part of the initial image and appreciate how our method successfully detects it.

\begin{table}
\begin{center}
\scalebox{1.1}{
\begin{tabular}{ |c|c|c|c| }
\hline
   & \textbf{Accuracy} & \textbf{FN Rate}  & \textbf{FP Rate} \\
   \hline
\textbf{Wagon 1} & 88.2  & 11.8 & 5.9 \\
\hline
\textbf{Wagon 2} & 100.0 & 0  & 14.71 \\
\hline
 & \textbf{94.1} & \textbf{5.9}  & \textbf{10.3} \\
\hline
\end{tabular} 
}
\end{center}
\caption{Full ID segmentation accuracy.}
\label{table:textacc_id}
\end{table}

\begin{table}
\begin{center}
\scalebox{1.1}{
\begin{tabular}{ |c|c|c|c| }
\hline
   & \textbf{Accuracy} & \textbf{FN Rate}  & \textbf{FP Rate} \\
\hline
\textbf{Wagon 1} & 93.4  & 6.6 & 0.7 \\
\hline
\textbf{Wagon 2} & 100.0 & 0  & 1.2 \\
\hline
 & \textbf{96.7} & \textbf{3.3} & \textbf{1.0} \\
\hline
\end{tabular}
}
\end{center}
\caption{ID characters segmentation accuracy.}
\label{table:textacc}
\end{table}

\subsection{Pantograph detection}
The pantograph was observed only in the afternoon test session, so for 18 (out of 34) sequences of the dataset. However, for each one of the 18 sequences the pantograph is correctly detected by the proposed solution.

\begin{figure}
\centering
\subfigure{\includegraphics[width=1\columnwidth]{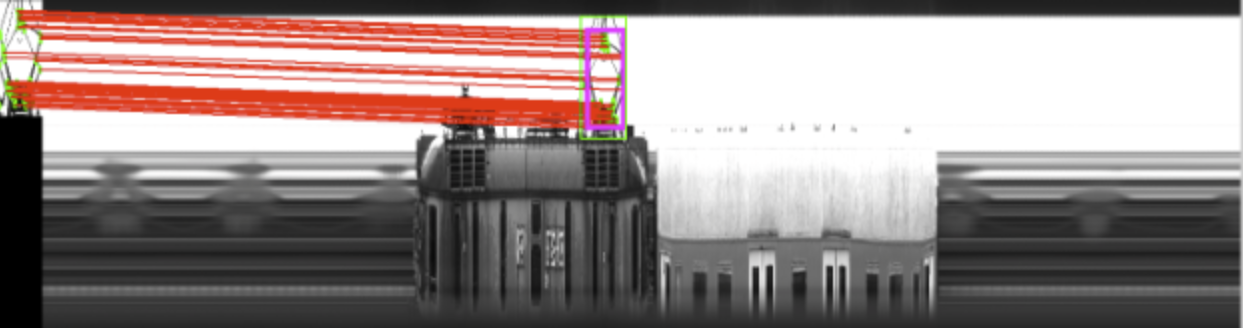}}
\subfigure{\includegraphics[width=1\columnwidth]{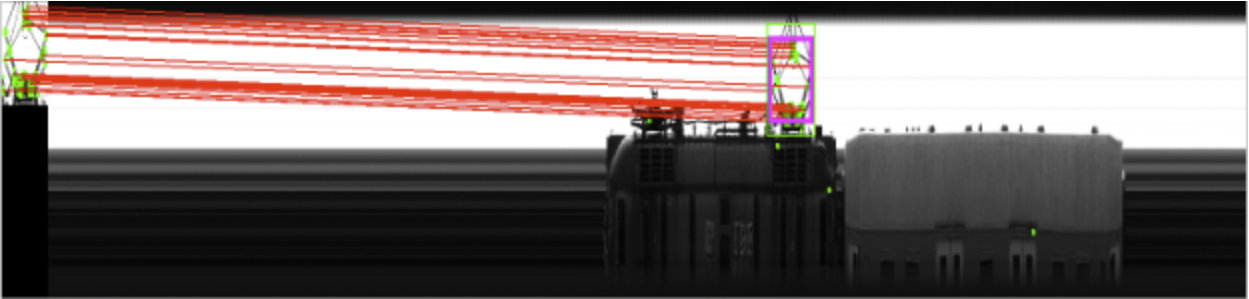}}
\subfigure{\includegraphics[width=1\columnwidth]{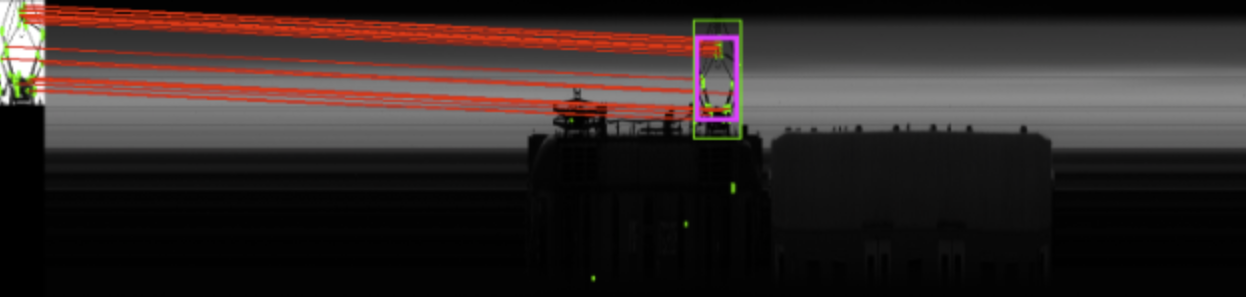}}
\caption{Examples of pantograph matching with different illumination conditions. The pantograph template is on the upper left side of each image.}
\label{fig:pantograph_samples}
\end{figure}

In \figurename~\ref{fig:pantograph_samples} we report some samples of the pantograph matching working under very different illumination conditions.

\subsection{Evaluation of the user interface}
Cognitive Walkthrough (CW)~\cite{wharton1994cognitive} is a usability inspection method whose objective is to identify usability problems, focusing on how easy it is for new users to accomplish predefined tasks. It is a technique that aims at detecting  errors in design that would interfere with the performance of users while using the interface. CW is usually carried out by specialists in the field of interface development and usability experts.  As the walkthrough proceeds, comments of the users are recorded.
We conducted an usability inspection of the proposed touch-based interface to assess its possible usability issues.
For this purpose we defined the following different tasks:
\begin{description}
  \item[T1] Load the most recent analysed sequence 
  \item[T2] Position the video on the sequence corresponding to the pantograph detection 
  \item[T3] Visualize areas of the wagon which temperature is higher than $50^{\circ}$ 
  \item[T4] Visualize the wagon ID number using the analysis results 
\end{description}
For each task we defined a sequence of actions with details about specific task flow from beginning to end. We asked 5 examiners to perform the defined task using the so-called \textit{think aloud} technique in order to record failure in the interaction and design suggestions.
Previous studies on usability testing~\cite{nielsen1993mathematical} showed that the number of usability problems $\mathbf{U_p}$  found in a usability test is:
\begin{equation}
\mathbf{U_p} = \mathbf{N} (1-(1- \mathbf{L} )^\mathbf{n} )
\end{equation}
where $\mathbf{n}$ is the number of users, $\mathbf{N}$ is the total number of usability problems in the design and $\mathbf{L}$ is the proportion of usability problems discovered while testing a single user. The typical value of $\mathbf{L}$ is 31\%, suggesting that 85\% of usability issues can be found with 5 testers. 

All the examiners were able to complete assigned tasks, reporting usability and user interface related problems while performing the evaluation.
Feedbacks from examiners allowed us to identify and correct minor but important usability issues, mostly regarding sizes and positions of objects on the screen or ambiguities in the use of textual labels.     

\section{Conclusions}

In this paper we introduced a multi-camera portal for train safety assessment. Our proposal is able to perform the analysis of multiple safety requirements of each train passing under the gantry without requiring the train to be stopped. 
We detailed the hardware used and the software developed to robustly acquire data from multiple high-rate sensors. Image processing and computer vision methods are applied on each data stream to extract meaningful information. We also presented our multitouch interface that enables an operator to quickly observe and simply interact with the processed data. The evaluation has shown the good performances of the analysis and the usability of the interface.

\section*{Acknowledgment}
This work was supported by the Integrated Intermodal System for Security and Signaling on Rail (SISSI) project, funded by Regione Toscana - Italy.

\bibliographystyle{spmpsci}      
\bibliography{SISSI_mva}   

\begin{thebibliography}{10}
\providecommand{\url}[1]{{#1}}
\providecommand{\urlprefix}{URL }
\expandafter\ifx\csname urlstyle\endcsname\relax
  \providecommand{\doi}[1]{DOI~\discretionary{}{}{}#1}\else
  \providecommand{\doi}{DOI~\discretionary{}{}{}\begingroup
  \urlstyle{rm}\Url}\fi

\bibitem{y2008methods}
Arcas, B.A.y.: Methods and apparatus for navigating an image (2008).
\newblock US Patent 7,375,732

\bibitem{FrenchTrainCrash}
BBC: Deadly french train crash at bretigny-sur-orge (2013).
\newblock \urlprefix\url{http://www.bbc.com/news/world-europe-23294630}

\bibitem{beck1973two}
Beck, F., Stumpe, B.: Two devices for operator interaction in the central
  control of the new cern accelerator.
\newblock Tech. rep., CERN (1973)

\bibitem{bjorneseth2012assessing}
Bj{\o}rneseth, F.B., Dunlop, M.D., Hornecker, E.: Assessing the effectiveness
  of direct gesture interaction for a safety critical maritime application.
\newblock International Journal of Human-Computer Studies \textbf{70}(10),
  729--745 (2012)

\bibitem{Brown03}
Brown, M., Lowe, D.G.: Recognising panoramas.
\newblock In: Proc. of International Conference on Computer Vision (ICCV),
  vol.~3, p. 1218 (2003)

\bibitem{camargo2011emerging}
Camargo, L.F.M., Edwards, J.R., Barkan, C.P.: Emerging condition monitoring
  technologies for railway track components and special trackwork.
\newblock In: 2011 Joint Rail Conference, pp. 151--158. American Society of
  Mechanical Engineers (2011)

\bibitem{Canny:1986:CAE:11274.11275}
Canny, J.: A computational approach to edge detection.
\newblock IEEE Trans. Pattern Anal. Mach. Intell. \textbf{8}(6), 679--698
  (1986).
\newblock \doi{10.1109/TPAMI.1986.4767851}.
\newblock \urlprefix\url{http://dx.doi.org/10.1109/TPAMI.1986.4767851}

\bibitem{delgado2014automatic}
Delgado, B., Tahboub, K., Delp, E.J.: Automatic detection of abnormal human
  events on train platforms.
\newblock In: IEEE National Aerospace and Electronics Conference, pp. 169--173
  (2014)

\bibitem{edwards2009advancements}
Edwards, J.R.: Advancements in railroad track inspection using machine-vision
  technology.
\newblock Ph.D. thesis, University of Illinois at Urbana-Champaign (2009)

\bibitem{Fischler:1981:RSC:358669.358692}
Fischler, M.A., Bolles, R.C.: Random sample consensus: A paradigm for model
  fitting with applications to image analysis and automated cartography.
\newblock Commun. ACM \textbf{24}(6), 381--395 (1981).
\newblock \doi{10.1145/358669.358692}.
\newblock \urlprefix\url{http://doi.acm.org/10.1145/358669.358692}

\bibitem{Forlines:2007:DVM:1240624.1240726}
Forlines, C., Wigdor, D., Shen, C., Balakrishnan, R.: Direct-touch vs. mouse
  input for tabletop displays.
\newblock In: Proceedings of the SIGCHI Conference on Human Factors in
  Computing Systems, CHI '07, pp. 647--656. ACM, New York, NY, USA (2007).
\newblock \doi{10.1145/1240624.1240726}.
\newblock \urlprefix\url{http://doi.acm.org/10.1145/1240624.1240726}

\bibitem{fumagallimultifunction}
Fumagalli, L., Tomassini, P., Zanatta, M., Libretti, G., Trebeschi, M.,
  Sansoni, G., Docchio, F.: Multifunction portals for train monitoring: Recent
  advances and innovative optoelectronic instrumentation.
\newblock In: X.~Perpinya (ed.) Reliability and Safety in Railway. InTech
  (2012).
\newblock \doi{10.5772/37626}

\bibitem{kin2009determining}
Kin, K., Agrawala, M., DeRose, T.: Determining the benefits of direct-touch,
  bimanual, and multifinger input on a multitouch workstation.
\newblock In: Proceedings of Graphics interface 2009, pp. 119--124. Canadian
  Information Processing Society (2009)

\bibitem{ViareggioIncident}
Landucci, G., Tugnoli, A., Busini, V., Derudi, M., Rota, R., Cozzani, V.: The
  viareggio \{LPG\} accident: Lessons learnt.
\newblock Journal of Loss Prevention in the Process Industries \textbf{24}(4),
  466 -- 476 (2011).
\newblock \doi{http://dx.doi.org/10.1016/j.jlp.2011.04.001}.
\newblock
  \urlprefix\url{http://www.sciencedirect.com/science/article/pii/S0950423011000362}

\bibitem{Lowe:2004:DIF:993451.996342}
Lowe, D.G.: Distinctive image features from scale-invariant keypoints.
\newblock Int. J. Comput. Vision \textbf{60}(2), 91--110 (2004).
\newblock \doi{10.1023/B:VISI.0000029664.99615.94}.
\newblock \urlprefix\url{http://dx.doi.org/10.1023/B:VISI.0000029664.99615.94}

\bibitem{muja_flann_2009}
Muja, M., Lowe, D.G.: Fast approximate nearest neighbors with automatic
  algorithm configuration.
\newblock In: International Conference on Computer Vision Theory and
  Application VISSAPP'09), pp. 331--340. INSTICC Press (2009)

\bibitem{nielsen1993mathematical}
Nielsen, J., Landauer, T.K.: A mathematical model of the finding of usability
  problems.
\newblock In: Proceedings of the INTERACT'93 and CHI'93 conference on Human
  factors in computing systems, pp. 206--213. ACM (1993)

\bibitem{1979:ots}
Otsu, N.: {A} {T}hreshold {S}election {M}ethod from {G}ray-level {H}istograms.
\newblock IEEE Transactions on Systems, Man and Cybernetics \textbf{9}(1),
  62--66 (1979).
\newblock \doi{10.1109/TSMC.1979.4310076}

\bibitem{pu2014study}
Pu, Y.R., Chen, L.W., Lee, S.H.: Study of moving obstacle detection at railway
  crossing by machine vision.
\newblock Information Technology Journal \textbf{13}(16), 2611--2618 (2014)

\bibitem{sacchipavisys}
Sacchi, M., Cagnoni, S., Spagnoletti, D., Ascari, L., Zunino, G., Piazzi, A.:
  Pavisys: A computer vision system for the inspection of locomotive
  pantographs.
\newblock In: Proc. of International conference on Pantograph Catenary
  Interaction Framework for Intelligent Control (2011)

\bibitem{schupfer2001fire}
Schupfer, H.: Fire disaster in the tunnel of the kitzsteinhorn funicular in
  kaprun on 11 nov. 2000.
\newblock In: Porc. of the Fourth International Conference on Safety in Road
  and Rail Tunnels (2001)

\bibitem{stelzer2014evaluating}
Stelzer, A., Sch{\"u}tz, I., Oetting, A.: Evaluating novel user interfaces in
  (safety critical) railway environments.
\newblock In: Human-Computer Interaction. Applications and Services, pp.
  502--512. Springer (2014)

\bibitem{thimbleby2007interaction}
Thimbleby, H.: Interaction walkthrough: evaluation of safety critical
  interactive systems.
\newblock In: Interactive Systems. Design, Specification, and Verification, pp.
  52--66. Springer (2007)

\bibitem{weichselbaum2013accurate}
Weichselbaum, J., Zinner, C., Gebauer, O., Pree, W.: Accurate 3d-vision-based
  obstacle detection for an autonomous train.
\newblock Computers in Industry \textbf{64}(9), 1209--1220 (2013)

\bibitem{wharton1994cognitive}
Wharton, C., Rieman, J., Lewis, C., Polson, P.: The cognitive walkthrough
  method: A practitioner's guide.
\newblock In: Usability inspection methods, pp. 105--140. John Wiley \& Sons,
  Inc. (1994)

\bibitem{zahler2008design}
Zahler, T.: A design process for constructing a user interface pattern library
  for touch-based applications in safety-critical environments.
\newblock In: System Safety, 2008 3rd IET International Conference on, pp.
  1--3. IET (2008)

\end{thebibliography}

\end{document}